%% file: 00_Frame.tex
\documentclass[conference]{IEEEtran}
\IEEEoverridecommandlockouts

\usepackage{cite}
\usepackage[utf8]{inputenc}
\usepackage{amsmath,amssymb,amsfonts}
\usepackage[vlined,ruled,linesnumbered]{algorithm2e} 
\usepackage[pdftex]{graphicx}
\usepackage{textcomp}
\usepackage{xcolor}
\usepackage{verbatim} 
\usepackage{paralist}
\usepackage{booktabs}
\usepackage{tabularx}
\usepackage{hyperref}

\DeclareMathOperator*{\argmin}{argmin}

\newcommand{\N}{\ensuremath{\mathbb{N}}}
\newcommand{\R}{\ensuremath{\mathbb{R}}}

\begin{document}

\input{10_Title}
\input{20_Introduction}
\input{30_RelatedWork}

\input{40_BigDataPlatform}

\input{60_UseCases}
\input{70_Conclusion}

\section*{Acknowledgment}
The work was supported by the German Federal Ministry of Education and Research (BMBF) under the project "KOARCH" (funding code: 13FH007IA6 and 13FH007IB6).

\bibliographystyle{IEEEtran}

\bibliography{%
literature%
}

\end{document}

%% file: 10_Title.tex
\title{Cognitive Capabilities for the CAAI in Cyber-Physical Production Systems}

\author{
\IEEEauthorblockN{Jan Strohschein\IEEEauthorrefmark{3}, Andreas Fischbach\IEEEauthorrefmark{2}, Andreas Bunte\IEEEauthorrefmark{1}, \\
Heide Faeskorn-Woyke\IEEEauthorrefmark{3}, Natalia Moriz\IEEEauthorrefmark{1} and Thomas Bartz-Beielstein\IEEEauthorrefmark{2}} \\
\IEEEauthorblockA{\IEEEauthorrefmark{1}OWL University of Applied Sciences and Arts, Institute Industrial IT, Lemgo, Germany\\
Email: andreas.bunte@th-owl.de, natalia.moriz@th-owl.de}
\IEEEauthorblockA{\IEEEauthorrefmark{2}TH Köln, Institute for Data Science, Engineering, and Analytics, Gummersbach, Germany\\
Email: andreas.fischbach@th-koeln.de, thomas.bartz-beielstein@th-koeln.de}
\IEEEauthorblockA{\IEEEauthorrefmark{3}TH Köln, Institute of Computer Science, Gummersbach, Germany\\
Email: jan.strohschein@th-koeln.de, heide.faeskorn-woyke@th-koeln.de}
}

\maketitle
\begin{abstract}
This paper presents the cognitive module of the cognitive architecture for artificial intelligence (CAAI) in cyber-physical production systems (CPPS). 
The goal of this architecture is to reduce the implementation effort of artificial intelligence (AI) algorithms in CPPS.
Declarative user goals and the provided algorithm-knowledge base allow the dynamic pipeline orchestration and configuration.
A big data platform (BDP) instantiates the pipelines and monitors the CPPS performance for further evaluation through the cognitive module. Thus, the cognitive module is able to select feasible and robust configurations for process pipelines in varying use cases.
Furthermore, it automatically adapts the models and algorithms based on model quality and resource consumption. 
The cognitive module also instantiates additional pipelines to test algorithms from different classes.
CAAI relies on well-defined interfaces to enable the integration of additional modules and reduce implementation effort.
Finally, an implementation based on Docker, Kubernetes, and Kafka for the virtualization and orchestration of the individual modules and as messaging-technology for module communication is used to evaluate a real-world use case.

\end{abstract}
\begin{IEEEkeywords}
Cognition, Industry 4.0, Big Data Platform, Machine Learning, CPPS
\end{IEEEkeywords}

%% file: 20_Introduction.tex
\section{Introduction}

Artificial Intelligence (AI) in Cyber-physical Production Systems (CPPS) can help to significantly reduce costs~\cite{Kagermann:2013}, but its implementation requires expert knowledge and thus might be cost-intensive~\cite{FSB20}.
To tackle these challenges, a modular and extendable Cognitive Architecture for Artificial Intelligence (CAAI) in CPPS was introduced in previous work~\cite{FSB20}.
The architecture defines process pipelines as a sequence of processing modules, e.g., a preprocessing module, followed by a modeling module wrapped up by an optimization module to find an optimal configuration of the model.
The usage of these cognitive capabilities for the selection of AI algorithms enable the system to learn over time and choose suitable algorithms automatically and thus replace the expert knowledge, to some extent.
This can be a key feature to reach a high degree of autonomy and efficiency in CPPS~\cite{BFS:2019}.

The work at hand describes the cognitive module and its functionality, which is further referred to as \textit{Cognition}, in more detail than~\cite{FSB20}. 
The task of the cognitive module is to propose candidate pipelines with proper parameters, compute several pipelines in parallel, evaluate the pipeline quality, and switch to promising pipelines during the operational phase, e.g., if they are likely more efficient w.r.t. accuracy or resource consumption.
Furthermore, the cognitive module does not require deeper AI knowledge from the user.
This paper focuses on the optimization use case, whereas the general concept will be extended to other use cases, such as Condition Monitoring, Predictive Maintenance, or Diagnosis.
The main contributions of this paper are:
\begin{itemize}
	\item Automatic algorithm selection and tuning in dynamic environments,
	\item automatic creation of machine learning pipelines, based on selected algorithms, and
	\item real-world evaluation of the cognitive module and available CAAI implementation for the use case on GitHub.
\end{itemize}

The remainder of this paper is organized as follows: Section~\ref{cha:relatedWorks} provides an overview of related works. 
The concept of the cognitive module is described in Section~\ref{cha:concept} along with components that are closely related and important for its behavior. 
Details about the implementation on the CAAI BDP and an evaluation on a real-world use case can be found in Section~\ref{cha:implementation}.
Finally, Section~\ref{cha:summary} discusses our major findings and resulting future research tasks.

%% file: 30_RelatedWork.tex
\section{Related Work}\label{cha:relatedWorks}
The main contributions of our work concern several research areas. Thus, the first  sub-section reviews the orchestration and scheduling of machine learning workloads on Kubernetes, an open-source system for automating deployment, scaling, and management of containerized applications. Details on Kubernetes can be found in \cite{Burn2019}. The second sub-section addresses the algorithm selection and tuning to generate feasible optimization pipelines.

\subsection{Orchestration and Scheduling on the Big Data Platform}
The CAAI uses Kubernetes for the orchestration of machine learning pipelines. 
Research found several projects that build machine learning workflows on top of Kubernetes.

Altintas et al., present Chase CI as a highly scalable infrastructure project for machine learning based on Kubernetes \cite{Alti2019}.
They build a cluster consisting of resources from 20 partner institutions and portray a deep learning use case where neural networks learn from weather data.
A list of steps is developed called "Process for the Practice of Data Science" to guide AI experts.
The work shows no signs of cognitive capabilities and is not easy to extend, as there are no templates or abstractions.
 
Subramaniam et al., propose abstractions for machine learning workloads in Kubernetes \cite{Subr2018}. 
They develop a set of Custom Resource Definitions (CRDs) and custom controllers for Kubernetes to make it easier for an AI expert to create new machine learning jobs on Kubernetes.
The work does not possess cognitive abilities or help the user to select a suitable algorithm.

A related project that stems from industry is Kubeflow.
It is meant to simplify the process of deploying machine learning workflows on Kubernetes and started as an internal Google TensorFlow framework that was open-sourced in late 2017. 
In Kubeflow a pipeline describes a single machine learning workflow, where each component is packaged as a Docker image. 
Those pipelines can be created programmatically with a domain-specific-language (DSL) they provide. 
It is also possible to convert Jupyter notebooks into a pipeline via a graphical-user-interface and a plugin \cite{Dell19}.
Even though the project provides a nice graphical interface it is targeted mostly towards data scientists and machine learning engineers. 
Right now the project allows to process data in batches and has no built-in streaming capabilities.
The user also has to build the pipeline by himself and Kubeflow does not support the user to choose the right algorithm for a use case.

Our cognitive module dynamically schedules pipelines on Kubernetes based on the available system resources. 
Several other works consider the scheduling of machine learning workloads on a Kubernetes cluster.
The researchers \cite{Kaur2019,Casq2019,Peng2018} use insights into cluster information and job states for dynamic allocation with a focus on energy efficiency and minimization of training times.
Peng et al., developed a schedular that is capable to update the resources based on training speed and predicted time needed for model convergence \cite{Peng2018}. 
The related works improve the performance compared to the standard Kubernetes scheduler or traditional cluster schedulers, e.g. Mesos and Yarn. 
They use the system resources efficiently but do not determine which jobs should be instantiated to solve a machine learning problem.
While those related works target machine learning on Kubernetes, we found no projects that specifically target machine learning for CPPS in Kubernetes.

\subsection{Algorithm Selection and Tuning}
Muñoz et al.~\cite{Mun15} provide a survey on methods to address the problem of algorithm selection for black-box continuous optimization problems.
They precisely describe the algorithm selection problem (ASP) as challenging due to the limited theoretical understanding of the strengths and weaknesses of most algorithms.
Furthermore, due to a large number of available algorithms, it is difficult or impossible to overview all algorithms.
A framework for ASP was introduced by Rice~\cite{Ric76} and described more recently by Smith-Miles~\cite{Smi09} as a four-step process.
However, the implementation is still challenging and often interpreted as a learning task~\cite{Smi09}.
Typically classification or regression models are employed to address this task~\cite{Mun15}.
Both methods come with some disadvantages, e.g., classification methods have to be re-trained if there are changes in the algorithm, whereas regression models are modular but with its higher number of elements, regression models are prone to failures~\cite{Mun15}.

One way to extract features of the problem instances is provided by Exploratory Landscape Analysis (ELA)~\cite{Mer11}.
ELA characterizes the problem by a larger number of numerical feature values that are grouped into categories.
These features are determined by numerical computations based on sampling of the decision space.
The advantage of these so-called \textit{low-level features} is, that they can be determined automatically, although they are related to some high-level features, whose creation requires knowledge.
Based on these features the selection of an optimizer process can be performed, e.g. by using a classifier~\cite{Mer11}.
A combined approach of Machine Learning (ML) and the application of ELA, where results from previous Black-Box-Optimization-Benchmarking workshops taking place at the Genetic and Evolutionary Computation Conference (GECCO) were used to train the selectors, can be found in~\cite{Ker19}. 
This approach currently comes with some drawbacks. 
Recent research empirically shows, that not all ELA features are invariant to rotation, translation (shifting), and scaling of the problem~\cite{Muno18a,Skvo20a}. 
Additionally, features are sensitive to the sampling strategy employed~\cite{Rena20a}. 

Automated Machine Learning (AutoML)~\cite{Feur15a, Feur20} and hyperheuristics can choose and configure a suitable algorithm automatically. 
That includes steps such as data pre-processing, algorithm selection, and hyperparameter optimization~\cite{Fusi18a, Olso16a, Thor13a}.
However, these approaches are built to process offline data, so they expect a training and a test data set.
In comparison to that, our approach has to deal with online data, so the data set is continuously growing and not available a priori.
Especially in the first few production cycles the amount of available data is not sufficient to partition the data set into test and training data as AutoML methods would require.

For the evaluation of the method performance on real-world problems, or, more precisely, real-world related test problems, Zaeffer et al. employed Gaussian process simulation for discrete optimization~\cite{Zaef17a}. 
This extends a premature work of varying Gaussian process model parameters within a certain range to retrieve instances from a problem class~\cite{Fisc16a}.
This enables the generalization of performance evaluation methods on problem classes, which is described in~\cite{Fisc18a}.
The application of Gaussian process simulation for the continuous domain can be found in~\cite{Zaef20a}.

The approach to tackle the ASP in this paper is based on the objective function simulation employing an initial design sampling of the process.
Conducting small-scale benchmark experiments determines feasible and well-performing algorithms from an available portfolio in an iterative process.

%% file: 40_BigDataPlatform.tex
\section{Concept}\label{cha:concept}
The concept of the cognitive module and its implementation is presented in this section.
In the first subsection, the general concept of the CAAI and the cognitive module is introduced.
Then a concept for defining declarative goals for the CAAI is presented.
In the following subsection, the description of algorithms is addressed, which is stored in the knowledge base and acts as a basis for the selection of algorithms.
A detailed description of the cognitive module follows.
Finally, the connection between the \textit{Cognition} and the other modules is described, since the \textit{Cognition} has to configure the different modules automatically.

\subsection{General concept of CAAI}
\begin{figure}[h]
\centering
\includegraphics[width=1\columnwidth]{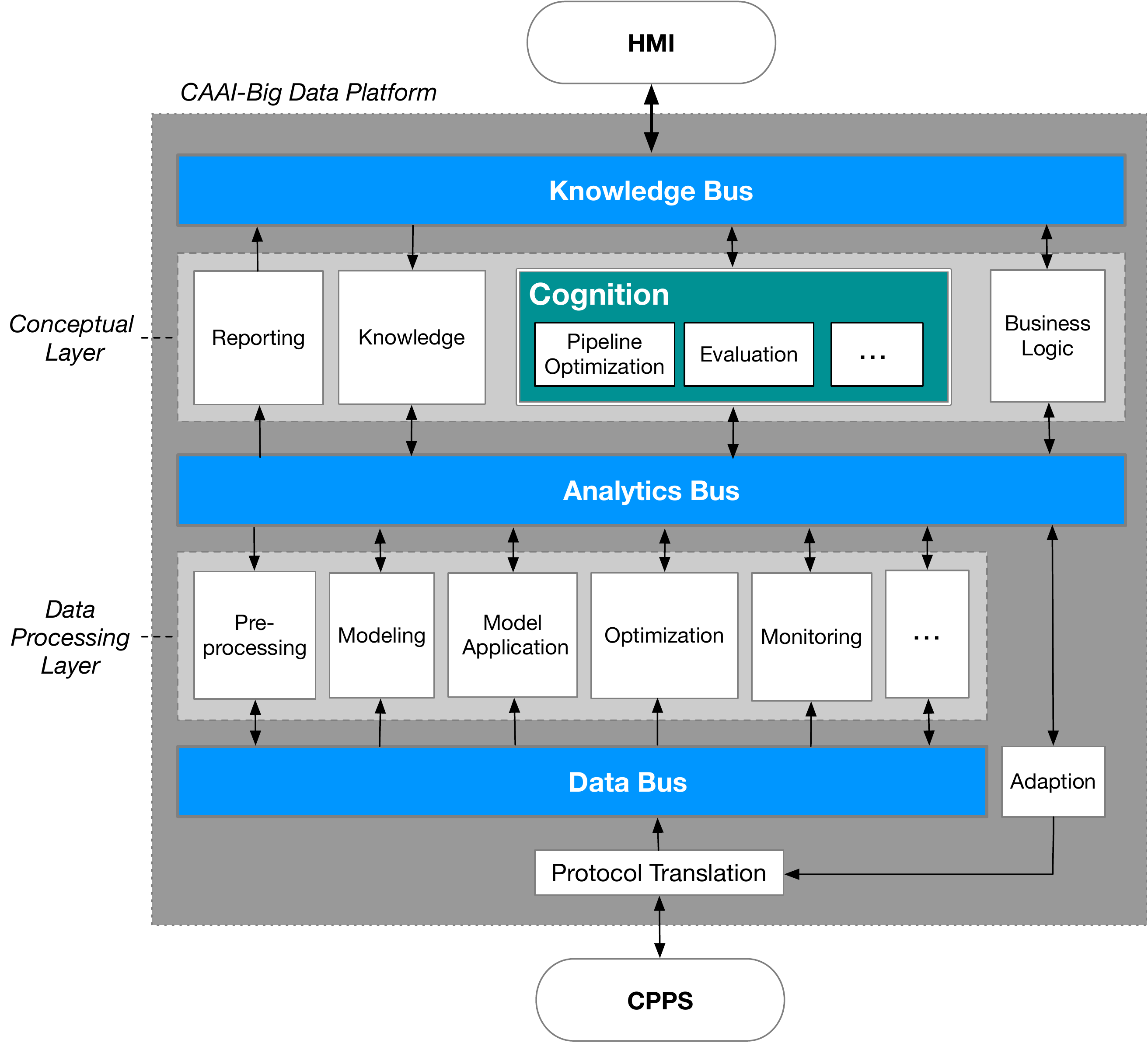}
\caption{CAAI Architecture, consisting of a CAAI-Big Data Platform, three bus systems, a conceptual layer, and a data processing layer.}
\label{fig:CAAI}
\end{figure}

The cognitive architecture CAAI builds upon the idea of modeling the information, data applications, and streams required for specific tasks in the I4.0 scenario while providing reliability, flexibility, generalizability, and adaptability. 
The concept, depicted in Figure~\ref{fig:CAAI}, is based on a three-tier architecture to simplify interoperability and ensure horizontal scalability.
The CAAI-BDP, depicted in dark grey, wraps the architecture and arranges software modules in two processing layers (shaded in light grey), the Data Processing Layer (DPL), and the Conceptual Layer (CL). 
The layers are connected via bus systems (data, analytics, and knowledge bus), which are colored in blue. 
Arrows demonstrate the designated information flow.

Data from a CPPS enters the system at the very bottom. 
The \textit{Protocol Translation} transforms incoming data and sends it to the data bus.
The pre-processing module receives the raw data, performs the necessary steps to clean the data, and publishes the results back to the data bus. 
Other modules in the DPL, such as \textit{Modeling} or \textit{Optimization}, utilize data from the data bus and transfer their analytical results onto the analytics bus. 
Modules in the CL process information about the user-defined aims and the business logic for a given use case. 
They evaluate the results from the analytics bus, determine the parameters to adjust the CPPS via the adaption module, and measure the overall system performance and available resources through the monitoring module. 
The CL modules also interact with the knowledge bus to generate reports for the user and to process new instructions. 
The human-machine-interface (HMI) communicates with the BDP through the knowledge bus, where the user can add new declarative goals during operation or adapt the knowledge base.
However, the central element of the architecture is the \textit{Cognition}, which selects, orchestrates, and evaluates different algorithms, depending on the use case. 
Therefore the composition of active modules and their communication over the bus system will change during run time. 
Providing a pre-defined set of modules and the capability to add new modules reduces the overall implementation complexity by building a cohesive yet modular solution. 
A more detailed description of the CAAI can be found in \cite{FSB20}.
In the work at hand, the \textit{Cognition} is described in detail. 
Furthermore, an implementation and evaluation of the cognitive module are performed in this paper.

\subsection{Declarative Goals}\label{sec:declarativeGoals}
To enable easy usage of the CAAI, the method to optimize the CPPS will be selected automatically. 
This is possible due to CAAI utilizing declarative goals, e.g, the user specifies the goal, not the single process steps to achieve it.
However, this declarative goal must be formulated at least once and goals such as \textit{Resource Optimization} are too unspecific and cannot be converted into an appropriate pipeline.
Thus, CAAI implemented a step-wise procedure to assist the practitioner in the process. 

The task of an algorithm in the context of an optimization problem is to find the setting of one or more control parameters $ x \in \mathbb{R}^n$ which minimizes (or maximizes) a function $y = f( x)$ subject to constraints $\Phi(x)$:
\begin{equation*}
\label{eq:argmin}
\begin{aligned}
	\argmin ~ & f( x) ~~ s.t. \\
	& \Phi( x)
\end{aligned}
\end{equation*}

\begin{figure}[h]
\centering
\includegraphics[width=0.6\columnwidth]{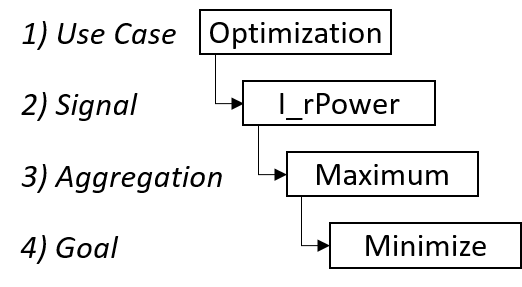}
\caption{Application of the four-stage declarative goal selection.}
\label{fig:fourstage}
\end{figure}

The formulation of an optimization problem typically involves three steps~\cite{Bhatti2000}:
\begin{enumerate}
	\item Selecting control variables $x$,
	\item choosing the objective function, and
	\item identify constraints on $x$.
\end{enumerate}
Inside CAAI the first and third step result in the definition of the parameters that control the process and adapt the CPPS via the adaption module~(see Fig.~\ref{fig:CAAI}). 
The \textit{Business Logic} observes the parameter constraints and verifies that all values are feasible before the \textit{Adaption} sends adjusted parameters to the CPPS.
For the second step, the formulation of the objective function, we developed a multi-stage goal selection, which guides machine operators through the goal selection.
This is presented through the four-stage selection for optimization.
In the first stage, the user selects the overall goal, such as optimization, anomaly detection, condition monitoring, or predictive maintenance.
In the second stage, the signals are selected. This could be all, many, or just a single signal, depending on the overall goal and the use case.
In the third stage, aggregation functions can be selected, such as mean, delta, min, or max value, or the value itself.
In the last stage, the user selects the optimization goal, e.g. minimizing or maximizing.
With this four-step process, the user can select the optimization goals on a more abstract level.
This is shown in Figure~\ref{fig:fourstage}, on an energy optimization use-case of a bakery, which is described in more detail in \cite{BFS:2019}.
The goal of the use-case is the minimization of the peak power consumption.
Therefore, the user chooses \textit{Optimization}, the \textit{power signal}, the \textit{maximum value} of the signal and \textit{minimize}.
As one can see, the usage of such applications is quite straightforward.

It is possible to select multiple overall goals, e.g. optimization and anomaly detection, where each goal results in a separate pipeline.
If the goal optimization is selected multiple times, a single criteria optimization algorithm is used in this work, since we do not have multi-criteria optimization algorithms in the portfolio until now.
The different optimization goals are normalized and charged with equal weights.
So, this approach enables the user to select also complex goals in an easy way.

\subsection{Algorithm Characteristics}
Algorithms that are available in CAAI are described in the knowledge module using a common schema to enable the cognitive module to select suitable algorithms. 
This schema was developed in this work, as shown in Table~\ref{tab:AlgorithmProperties}.
The presented properties can be divided into mandatory properties that must have a certain value to enable algorithm usage, such as the property \textit{reach aim}.
Other properties represent just the performance of the algorithms and thus the best matching algorithms can be chosen, such as the property \textit{Performance}.
The required input data, output data, and the aim that should be reached are mandatory properties and enable to determine which algorithms can be used for a certain pipeline.

The property \textit{algorithm class} is used as a high-level property for the selection process.
According to the \textit{No free lunch theorem}~\cite{Wolp97a} we assume, that there is no single optimizer, which performs best on all optimization problems.
Therefore, the CAAI provides several optimizers from different optimization algorithm classes with different features, operators, and parameters.
Most likely algorithms from different classes will perform differently on problems according to both, the best-found value and also resource consumption. 
It might be beneficial, to ensure the test of a broad range of algorithms w.r.t. the available resources.
Furthermore, it can be learned that a certain class of algorithms is suitable or unsuitable for a certain CPPS.
The listed algorithm classes (see Table~\ref{tab:AlgorithmProperties}) are selected based on a taxonomy by Stork et al.~\cite{Stor18a}.
In an increasing order the given classes \textit{Hill-climber}, \textit{Trajectory}, \textit{Population}, and \textit{Surrogates} arguably reflect the complexity of its members. 
\begin{itemize}
	\item Hill-climbers are more or less greedy algorithms, especially useful for local search and often used in combination with more sophisticated algorithms (e.g., Surrogate-based optimizers). 
	\item Trajectory algorithms are able to escape local optima to avoid premature convergence. 
	\item Population-based algorithms imitate evolutionary development of populations of individuals (candidate solutions) over several generations and can solve complex and difficult optimization problems, although they are sensitive to their parameters (e.g., the population size, crossover and mutation rate, number of iterations). 
	\item Surrogate-based optimizers in our approach use a surrogate model (e.g., a regression model) as a replacement for the (potentially costly) objective function to estimate the objective function (via model predictions) relatively cheap (in comparison to the objective function itself). 
	Then the surrogate model is searched for the optimum and the found candidate(s) are evaluated on the objective function. 
	The retrieved values are used to update the model in the next iteration.
\end{itemize}
As the application of the CAAI tries to minimize the required knowledge regarding data science and algorithmic concerns, and the occurring use cases, i.e., the problem class and its features can be considered a priori unknown, it is beneficial to have several different algorithms with a broad variety of implemented concepts available. 
This should increase the possibilities of an algorithm selection component like the cognitive module to select an efficient algorithm according to the problem.
Three properties are specified for a certain use case, i.e., \textit{Performance}, \textit{computational effort}, and \textit{RAM usage}, and are updated by the \textit{Cognition} during the evaluation process.
An algorithm may not be suitable for a specific use case, even if the properties indicate a good match.
However, with the dynamic parameters, the CAAI is able to overcome this challenge and skip those algorithms.

\begin{table}[htb]
\caption{Properties of Algorithms, italic properties can be updated by the cognition module.}
\centering 
\begin{tabular}{l|l }  
Property & Values \\
\hline
Input data & Continuous, discrete, hybrid, timed \\ &Automata, 
 neuronal net,...\\
Output data & timed Automata, neuronal net, (discrete, \\
& continuous or time) anomaly,...\\
Reach aim & Optimization (min/max), CM, Anomaly \\ & detection, Diagnosis\\
Algorithm class & Hill-climber, Trajectory, Population, Surrogates\\
\hline
Use multithreads & true, false\\
Min Training Data & 0..n\\
Prefer usage & true, false \\
Avoid usage & true, false\\
\hline
\textit{Performance} & 0..1\\
\textit{Computational effort} & 0..1\\ 
\textit{RAM usage} & 0..1 
\label{tab:AlgorithmProperties} 
\end{tabular}
\end{table}

\subsection{Cognitive Module}
The purpose of the \textit{Cognition} is to select and parameterize suitable algorithms for a given use case.
The selection is based on four kinds of information: \textit{(i)} the use case and knowledge about suitable algorithms to reach the goal; \textit{(ii)} specifications of the data, such as available data, and type of data; \textit{(iii)} characteristics of the algorithms, such as run time or memory consumption; \textit{(iv)} experience about the performance of the algorithms in previous applications.

The problem is given by the user, as described in Section~\ref{sec:declarativeGoals}.
Information about the algorithms is stored in the knowledge module. 
Every algorithm has a signature, that describes the data required by the algorithms and the data they provide. 
Based on this information possible pipelines can be identified and rated regarding the expectations to be suitable to solve the problem.
The rating is performed based on the three characteristics mentioned above (\textit{(ii) -- (iv)}).

Typically algorithms have some requirements, e.g., regarding the number of needed training data, or data types.
These are hard criteria, which do not allow the usage of the algorithm for the type of system if they are not fulfilled.
However, since the amount of data is continuously growing during production cycles, the algorithm may become feasible at a later point in time. 
These criteria are checked frequently by the cognitive module. 
If the performance is dramatically decreasing or stagnating over a number of production cycles, perhaps due to premature convergence of an algorithm, the selected pipeline can be changed in between.

Depending on the available resources, the dimension of the data, and the amount of available data, the computational effort of each pipeline can be rated. 
Since resources are limited it is required to consider whether one pipeline with expected good results but high resource usage or several pipelines with a lower chance of good results should be evaluated.
In an extreme case, it might be impossible to apply a certain pipeline, if the resources are very limited.
So the resources are major criteria for the selection of suitable pipelines.

To rate the quality of results for a certain pipeline, experiences from previous experiments can be used.
Therefore, the quality of each experiment is evaluated and the related quality parameter in the knowledge base is adapted.
In this first stage of development, the findings from previous experiments can only be selected from the same type of machine, since it is not exactly known which characteristics influence the quality of results.
However, it supports the efficient selection of suitable algorithms and enables the transfer of learning results regarding the same machine types.

\begin{algorithm}[!htb]
	\SetAlgorithmName{Algorithm}{} \\ 
	\KwIn{Initial design size $s$, selection step size $\theta$, algorithm characteristics $KB$, goal $g$, available resources $r$, historical data $d_H$}

	define List for evaluation $e$ \\
	define List data $d$ \\ 
	define List pipeline resource consumptions $p_r$ \\ 
	define Parameter $x$ \\	
	\If{($|d_H|$ = 0)} {
 		List parameter $l \gets$ createInitialDesign(s, KB) \\
 		\ForAll{(parameter l)} {
			$d \gets d$ + applyToCPPS(l) \\
			$x = l$ \\
 		}
	} \Else {
		$d,x \gets$ historicalData $d_H$
	}
 \Repeat{true} {
	\If{(nrIterations $\%\, \theta = 0 \vee \zeta = 1$)} {
	$\zeta =0$\\
		testFunctionSet $S \gets$ generateTestFunctions(d) \\
	
		List $p \gets$ determineFeasiblePipelines($KB, g, d$) \\
		$p \gets$ selectCandidatePipelines($KB, r, e$) \\

		\ForAll(in parallel){$p[i]$} {
			// perform tuning and benchmarking \\
	 		$e \gets$ applyPipeline($p[i]$, S) \\ 
	 	}		
		$KB, p_{best} \gets$ ratePipelines($KB, p, e$) \\
	}
	$x_{best} \gets$ getBestX($p_{best}, d, x$) \\ 
	
	\If{($|x-x_{best}| \geq  \epsilon$)} {
		$e \gets$ applyToCPPS($x_{best}$) \\
		$x = x_{best}$
	}
	$d \gets$ d + receiveNewData()\\
	\If{( since $\theta/2$ steps stagnation $\vee$ performance decrease)} {
		$\zeta = 1$
	}
	
}
\caption{Cognition Module for Optimization}
\label{algo:Cognition}
\end{algorithm}

Algorithm~\ref{algo:Cognition} provides detailed insights into the cognition module behavior for optimization use cases.
At first, needed variables are defined (lines 1-4).
If no historical data are available, an initial design representing a list of different parameter configurations is created, the parameters are applied to the CPPS and the resulting data are stored in the list $d$. 
As several design methods exist and due to the modular design of the architecture, the design method can easily be adapted for special purposes of the CPPS.
Currently, a full factorial design is the default configuration, and we recommend space-filling designs.
For several design methods and corresponding optimality criteria, we refer to~\cite{Atki12a,Huss11a,Morr95a}.

After the initialization phase, the process of the algorithm selection based on data-driven simulation and benchmarking is started within an infinite loop (line 12).
Since we do not want to change pipelines in every iteration, a user-defined step size $\theta$ and a variable $\zeta$ are checked, to potentially change he pipeline only after each $\theta$ steps or in situations of prolonged stagnation or performance decrease.
Then, a test function set $s$ is generated based on the data $d$ (line 15). 
Gaussian process simulation is employed for this task~\cite{Zaef20a}.
Simulation intends to reproduce the covariance structure of a set of samples (in this case, the process data gathered with the initial design), which maintains the topology of the problem landscape.
The intention is to analyze the behavior of the performance for candidate algorithms on problems similar to the CPPS problem
The generation of several different instances, either via Spectral simulation or simulation by decomposition (see~\cite{Zaef20a} for details), allows a benchmarking of potentially feasible algorithms.
With this benchmark, the algorithms can be analyzed regarding their resource consumption (computation time and memory consumption) and performance (based on their rank, not the achieved values, as the problem instances most likely are different in that regard) even for a larger number of production cycles. 
We assume, that the resource consumption depends on the current machine load, the number of function evaluations to perform, and the dimensionality of the problem, but not on the problem landscape structure. 
Consequently, the resource consumption can be analyzed quite accurately using simulations.

The algorithm description from the knowledge base is used to determine a list of all feasible pipelines in line 16.
Based on the available resources, the knowledge base, and possibly existing earlier evaluations, the most suitable pipelines are selected in line 17.
This ensures, that pipelines can be excluded, which are already known to not compute a result in-time. 
These pipelines are applied in parallel to the test function set $s$, which is used for parameter tuning and benchmarking of the pipelines (line 18-20).
Instances are drawn randomly for each step, tuning the algorithms first with an equal budget, and benchmarking on a different instance afterwards. 

The final decision, which pipeline is used, is described in line 21.
As the overall performance measure, a weighted normalized aggregation of the achieved performance on the simulation instance, the memory consumption, and the used CPU time is computed and assigned to each pipeline. 
For an overview and discussion of issues and best-practices of benchmarking in general, and performance assessment in particular, we refer to~\cite{Bart20garxiv}.
A normalized processing time, computed by dividing the used CPU time by the runtime of a standard algorithm, is suggested by Johnson and McGeoch~\cite{John07a} and Weise et al.~\cite{Weis14a}.
Inspired by this idea, we chose to consider the baseline comparator, i.e., the random search, as the reference algorithm. 
The goal is, to reward efficient usage of resources, leading to higher accuracy compared to the baseline and relative to the competing algorithms.
On the other hand, algorithms that do not spend too many resources, due to their simplicity, should not be overrated, when they do not achieve proper objective function values. 
Otherwise they might be good fallback choices, when system load is high and few resources are available, at least if they perform better than the baseline. 
Consequently, algorithms performing worse than the baseline, will be removed from this iteration. 

The best pipeline in the remaining list is chosen for application on the CPPS. 
If the list is empty, the current $x$ will be returned.
Should the new $x_{best}$ differ significantly from the current $x$, the new $x_{best}$ is applied to the CPPS for the next iteration (line 23-25).

The new data produced by the CPPS are added to $d$ (line 26).
If there is no significant improvement after half of the step size, the $\zeta$ is set to 1, to perform an unscheduled algorithm selection cycle. 
This should help at the beginning of the process to recover poor decisions.

%% file: 60_UseCases.tex
\section{Implementation and Evaluation}\label{cha:implementation}
This section presents the implementation and integration of the \textit{Cognition} on the BDP. Subsequently, it describes the real-world evaluation use case and the obtained results.
The implementation, which is described in this section, is available on GitHub~\footnote{\url{https://github.com/janstrohschein/KOARCH/tree/master/Use\_Cases/VPS\_Popcorn\_Production/Kubernetes}}.

\subsection{Big Data Plattform}
The BDP runs on Kubernetes, which orchestrates the different services.
Kubernetes enables a declarative cluster management, e.g. the user submits a specification of the desired application state to Kubernetes and its controller takes the required measures to reach this state.
The user composes the application of different building blocks, which fits the modular approach of the CAAI architecture very well.
The smallest building block in Kubernetes is called a pod and consists of one or more application containers and optionally a volume for data storage.
The BDP uses two higher-level Kubernetes objects for most of the services, namely the deployment and the job.

\begin{figure}[h]
	\centering
	\includegraphics[width=1\columnwidth]{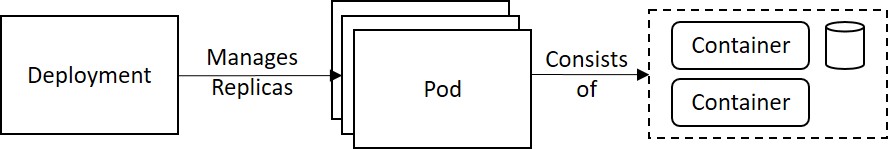}
	\caption{A deployment defines the lifecycle of a pod, which consists of one or more containers and an optional volume for information storage. Submitting a deployment to the controller instructs Kubernetes to instantiate the pod on a node and ensure its continuous availability in the cluster.}
	\label{fig:deployment_overview}
\end{figure}

The deployment is used for all modules of the BDP that require continuous operation, e.g. the cognitive module, the messaging solution or also the container registry. 
A deployment, as shown in Figure~\ref{fig:deployment_overview}, specifies the number of pod replicas, which the Kubernetes controller instantiates and monitors on the available nodes in the cluster. 
The controller will start new instances if a single pod or a complete node fails to get the cluster back to the desired deployment state.\\

\begin{figure}[h]
	\centering
	\includegraphics[width=1\columnwidth]{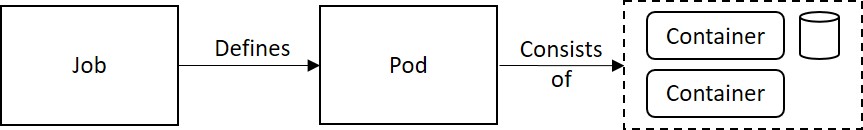}
	\caption{A job defines a pod, which consists of one or more containers and an optional volume for information storage. The job specifies a workload to process with a desired number of completions or parallelism, to run the same algorithm multiple times or launch several pods to work on one job in parallel.}
	\label{fig:job_overview}
\end{figure}

The job is meant for one-off execution of a task, e.g. a part of a data processing pipeline.
A job, as seen in Figure~\ref{fig:job_overview}, defines a pod and the desired amount of parallelism or the number of allowed retries, if the job fails during execution.
The Kubernetes controller will track the job progress and manage the whole lifecycle of the job to free up resources after completion. 

\begin{figure}[h]
	\centering
	\includegraphics[width=1\columnwidth]{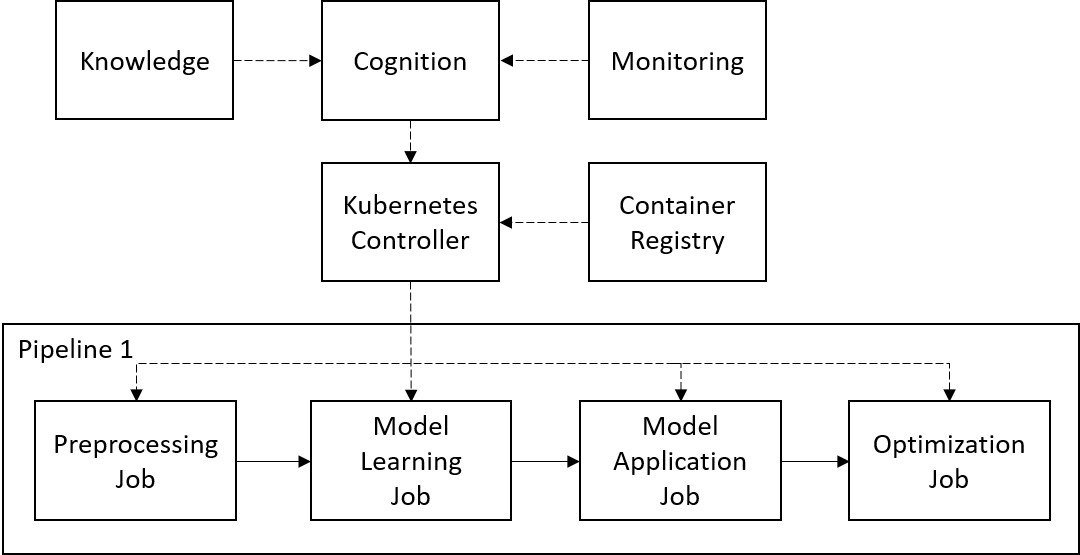}
	\caption{The \textit{Cognition} creates new data processing pipelines based on information about the available algorithms from the knowledge module and information on current resource usage provided by the monitoring module. The \textit{Cognition} decides on one or more pipelines and instructs the Kubernetes Controller to instantiate the data processing modules for each. The Kubernetes Controller loads the container images from the registry and starts all the jobs that form the data processing pipeline.}
	\label{fig:cognition_creates_pipeline}
\end{figure}

The complete process to dynamically create a data processing pipeline is depicted in Figure~\ref{fig:cognition_creates_pipeline}. 
The \textit{Cognition} decides which algorithms should be tested on the current use case based on information about available cluster resources from the monitoring module and the knowledge on available algorithms and their properties.
The cognitive module can specify and submit new jobs to dynamically build data processing pipelines.
The controller subsequently pulls the container images for the given jobs from the container registry and instantiates them.

Kubernetes defines all resources declaratively via the YAML Ain't Markup Language (YAML) and Listing YAML~\ref{algo:job} shows an exemplary job definition.
Line 1-2 define the type of Kubernetes resource, while line 3-4 specify job metadata, e.g., assign the algorithm name as an identifier for the job, and line 5 and onward describe the job at hand.
Lines 6-8 specify the Kubernetes configuration for the job, i.e., the number of retries (line 6), the maximum allowed time for processing (line 7), and the time before a finished job will be deleted by the controller (line 8).
The actual job can consist of one or more containers and the associated configuration is shown in lines 9-15.
A job template contains the container image (line 13) and the command to execute within the container on start-up (line 14), e.g., execute the Random forest implementation with Python. 
It is possible for the cognition to adjust the arguments to pass into the container (line 15), e.g. how many trees the Random forest algorithm will create or which is the criterion for a split.
Thus, the \textit{Cognition} can dynamically configure and instantiate a job that applies the Random forest algorithm on the production data.

\setcounter{algocf}{0}
\begin{algorithm}[!htb]
	\SetInd{0.3em}{0.5em}
	\SetAlgorithmName{YAML}{} \\ 
	apiVersion: batch/v1\\
	kind: Job\\
	
	\SetKwBlock{Fna}{\textnormal{metadata: }}{}
	\Fna{
	name: Random\_Forest
	}
	\SetKwBlock{Fnb}{\textnormal{spec: }}{}
	\Fnb{
	backoffLimit: 5\\
	activeDeadlineSeconds: 20\\
	ttlSecondsAfterFinished: 60\\
	\SetKwBlock{Fnc}{\textnormal{template: }}{}
	\Fnc{
	spec:\\
	containers:\\
	\SetKwBlock{Fnd}{\textnormal{-name: random\_forest}}{}
	\Fnd{
	image: caai/random\_forest\\
	command: ["python", "-u", "random\_forest.py"]\\
	args: ["NumberOfTrees=5"]\\
	}	
	restartPolicy: OnFailure
	}
	}

	\caption{Kubernetes Job Definition}
	\label{algo:job}
\end{algorithm}

The \textit{Cognition} uses the knowledge base to compose feasible pipelines and configure the available algorithms.
In this work, we used the YAML-format also for the hierarchical knowledge representation as a simple and straightforward way to represent algorithm knowledge that fulfills the requirements for extensibility and modularity.
The representation structure is based on the 4-step process to define declarative goals, presented earlier in Fig.~\ref{fig:fourstage}.
Listing YAML~\ref{algo:knowledge} shows the examplary knowledge representation of the Random forest algorithm.
Lines 1-3 represent the declarative user goals and the following lines describe the algorithms that are suited to achieve the individual goals.
Algorithms are described with the parameters, metadata about the algorithm, and the required input.
Parameters (line 6-11) are described by the type, minimum, maximum, and default value of the parameter.
The metadata (line 12-20) is a representation of Table~\ref{tab:AlgorithmProperties}, for a certain algorithm.
Not initialized values are indicated by -1. 
Those values are determined during the run time of the algorithms through the Cognition.
The \textit{efficiency} is calculated based on the model quality and the utilized resources as a selection criterion for the \textit{Cognition}. 
The input (line 21) for the algorithm is described as a string, e.g. preprocessed data.
All algorithms that provide this type of input are candidates for a pipeline and are described on the next hierarchy level.
The \textit{Cognition} repeats this selection process until the required input is designated as raw data which marks the end of a pipeline.
Thus, the representation enables a fast and easy composition of feasible pipelines and provides all necessary information for the \textit{Cognition} to evaluate those pipelines and decide which is the best suited for the production process.

\begin{algorithm}[!htb]
	\SetInd{0.3em}{0.5em}
	\SetAlgorithmName{YAML}{} \\ 
	\SetKwBlock{Fna}{\textnormal{Optimization: }}{}
	\Fna{
	\SetKwBlock{Fnb}{\textnormal{minimize: }}{}
	\Fnb{
	\SetKwBlock{Fnc}{\textnormal{Minimum: }}{}
	\Fnc{
	\SetKwBlock{Fnd}{\textnormal{Algorithms: }}{}
	\Fnd{
	\SetKwBlock{Fne}{\textnormal{Random Forest: }}{}
	\Fne{
	\SetKwBlock{Fnf}{\textnormal{parameter: }}{}
	\Fnf{
	\SetKwBlock{Fng}{\textnormal{NumberOfTrees: }}{}
	\Fng{
	type: int\\
	default: 4\\
	min: 1\\
	max: 100\\
	}
	}
	\SetKwBlock{Fnh}{\textnormal{metadata: }}{}
	\Fnh{
	Class: Surrogate\\
	Performance: -1\\
	Computational Effort: -1\\
	RAM usage: -1\\
	Min training data: 5\\
	}
	input: preprocessed data\\
	}
	}
	}
	}
	}

	\caption{Knowledge Representation}
	\label{algo:knowledge}
\end{algorithm}

\subsection{VPS Use Case}
We use the versatile production system (VPS), which is located in the SmartFactoryOWL, for evaluation of the cognitive module.
The VPS is a modular production system, which processes corn to produce popcorn which is used as packaging material.
Due to its modularity, it can be adaptd to the current order easily. 
Efficiently operating the VPS is a challenge because of many parameters influence the result, which cannot be measured inline, e.g., the moisture of the corn.
More details about the use case can be found in \cite{FSB20}.
Thus, a data-driven optimization is a promising method to increase efficiency, which is performed using the CAAI and the introduced cognitive module.

The goal is to operate the VPS at or near the optimum. 
This is difficult because the quality of the corn has a large influence on the optimal process parameter.
Depending on the quality the change of volume between the corn and the popcorn differs in a large range and can not be measured beforehand.
Thus, an optimization during the plants' run time is necessary.

In this paper, we take a typical application, where the corn is delivered in batches.
Since the quality between batches can differ, optimization is performed for each batch.
The batches are optimized independently, so parameters of the production plant, such as the size of a portion, can be changed between batches.
This describes the typical usage of the plant. 

The amount of corn that is filled into the reactor has to be optimized, to get the required amount of popcorn.
The overage of popcorn produced in one batch, or not fully filled boxes cannot be used, so it is wasted.
The optimum is a trade-off between three minimization functions: the energy consumption ($f_1$), the processing time ($f_2$), and the amount of corn needed for a small box ($f_3$).
These functions are conflicting to some degree. 
The optimization result is a parameter value $x$ for the dosing unit that indicates the runtime of the conveyer and thus influences the amount of corn.
As the given optimization problem can be regarded as relatively straightforward, we will apply a single objective optimization algorithm and compute a weighted sum of the objectives.
This results in the following optimization problem:
\begin{equation}
\label{eq:prob}
\min ~\sum_{i=1}^{3} w_i f_i(x); \text{~~w.r.t~~}  w_i > 0 \text{ and } \sum_{i=1}^{3}w_i = 1
\end{equation}
The scalar weights of the corresponding objectives, $w_i$,  are chosen based on user's preferences. 
As a default, equal weights are used. 

\subsection{Results}
To evaluate the cognitive module, data from the real-world VPS was acquired. 
A set of 12 different settings for the runtime of the conveyer was used, each repeated three times which results in 36 production cycles in total.
This data was used to fit a Gaussian process model. 
To retrieve an accurate simulation of the popcorn production, a conditional simulation was used (hereafter referred to as ground-truth), so consequently, the model respects the data points. 
Whereas the test instances, which will be used to perform the benchmark experiments, will use unconditional simulation (hereafter referred to as simulations).

\begin{figure}[!tb]
	\includegraphics[width=\columnwidth]{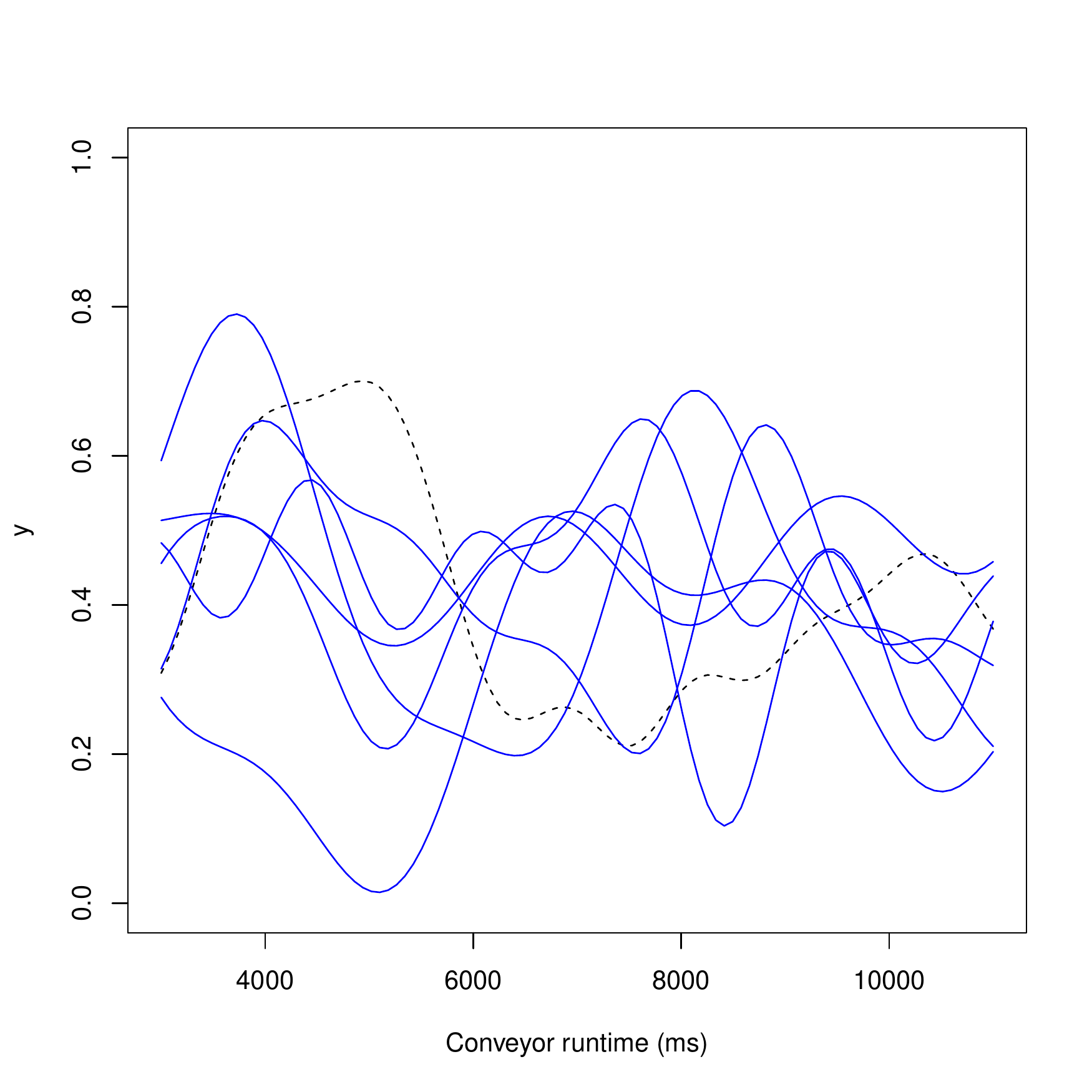}
	\caption{This plot shows the resulting VPS problem instances based on the real-world data taken from the machine. 
	The dashed line shows the conditional simulation, and the solid blue curves represent the unconditional simulations used as test instances for further benchmarking of the algorithm pipelines. 
	The x-axis shows the runtime of the conveyor in ms, and the y-axis shows the equally weighted normalized aggregated objective function value of the objectives process time, amount of corn, and energy consumption. }
	\label{fig:simulations}
\end{figure}

The resulting problem instances are shown in Fig.~\ref{fig:simulations}. 
It can be seen, that the simulation instances (shown as solid blue lines) arguably reflect the character of the ground-truth (the dashed line), but clearly have different forms of peaks and valleys. 

\begin{table}[!tb]
\caption[Optimizers parameter settings]{Settings of parameter ranges and corresponding default values of chosen optimizers}
\label{tab:algorithms}
\begin{tabularx}{\linewidth}{lXrl}
  \toprule
Parameter & Range & Default & Family \\
  \midrule\midrule
  \textbf{Differential evolution} & & & Population \\
  \midrule
popsize & $\N_+$ & $5$ &  \\ 
strategy & $\{1, 2, 3, 4, 5\}$ & $2$ & \\ 
F & ${[}0, 2{]}$ & $0.8$ & \\ 
CR & ${[}0, 1{]}$ & $0.5$ & \\ 
c & ${[}0, 1{]}$ & $0.5$ & \\ \midrule 
  \textbf{Generalized SA} & & & Trajectory \\
  \midrule
temp & $\N_+$ & $100$ & \\ 
$q_v$ & $\R$ & $2.5$ & \\ 
$q_a$ & $\R$  & $-1$ & \\ \midrule 
  \textbf{Kriging (SPOT)} & & & Surrogate \\
  \midrule
 designSize & $\N_+$ & $7$ & \\ 
 designType & $\{Lhd, Uniform\}$ & $Lhd$ & \\ \midrule 
  \textbf{Random forest (SPOT)} & & & Surrogate \\
  \midrule
 nrTrees & $\N_+$ & $500$ & \\ 
 designSize & $\N_+$ & $7$ & \\ 
 designType & $\{Lhd, Uniform\}$ & $Lhd$ & \\ \midrule 
  \textbf{L-BFGS-B} & & & Hill-climber \\
  \midrule 
   lmm & $\N_+$ & $5$ & \\ 
  \midrule

 \textbf{Uniform random sampling} & & & Baseline \\
 \bottomrule
\end{tabularx}
\end{table}

R version 3.6.3 was the used software platform~\cite{R20a} to perform the experiments.
To estimate the consumption of CPU time per optimizer, the \textit{tictoc} package in version 1.0, and for the monitoring of the memory consumption the \textit{profmem} package in version 0.5.0 are used.  
The Gaussian process simulations for continuous problem spaces are computed using the \textit{COBBS} package in version 1.0.0, available on GitHub~\footnote{\url{https://github.com/martinzaefferer/COBBS}}.
We use the Differential evolution implementation in the \textit{DEoptim} package in version 2.2-5, the Generalized simulated annealing (Generalized SA) implementation in the \textit{GenSA} package in version 1.1.7, the Kriging and Random forest based optimization from the \textit{SPOT} package version 2.0.6, and the L-BFGS-B implementation from the \textit{stats} package included in the R version. 
The considered algorithms with chosen parameters and corresponding values are summarized in Table~\ref{tab:algorithms}.
As a detailed description of the algorithms is out of scope of this paper, we refer to related publications.
Uniform random sampling does not belong to any of the before mentioned algorithm families, but is added as a baseline comparator and works without any control parameter. 
The L-BFGS-B, a limited memory variant respecting bound constraints of the \textit{Broyden-Fletcher-Goldfarb-Shanno} (BFGS) algorithm, is controlled by the parameter lmm, which sets the number of BFGS updates~\cite{Byrd95a}.
The Differential evolution algorithm is mainly controlled by the number of individuals per generation (popsize), the evolution strategy applied (strategy), the stepsize (F), the crossover probability (CR), and the crossover speed adaptation (c), for details we refer to~\cite{Mull11a}.
Generalized SA is a variant of the famous Simulated annealing global optimization algorithm, mimicking the cool-down process in metallurgy~\cite{Xian13a}.
The parameter temp initializes the starting temperature, $q_v$ and $q_a$ are parameters for the visiting and acceptance distribution respectively, controlling the candidate solutions and the acceptance of candidates with worse performance values to potentially escape local optima. 
Details about SPOT can be found in~\cite{Bart17parxiv}.

Please note that some values were chosen according to the relatively low budget of function evaluations, which is suitable for the given use case.  
This is especially the case for parameters that, e.g., control the number of candidate solutions per iteration. 
See for example, the designSize, which is the size of the initial design of both chosen surrogate-model based optimizers, or the popsize, which is the number of individual solutions for each iteration of the Differential evolution algorithm.

At first, the applicability of the simulations for the VPS problem is analyzed experimentally.
We assume a correlation between the performance of the available algorithms on the ground-truth and the simulations. 

\begin{figure}
	\includegraphics[width=\columnwidth]{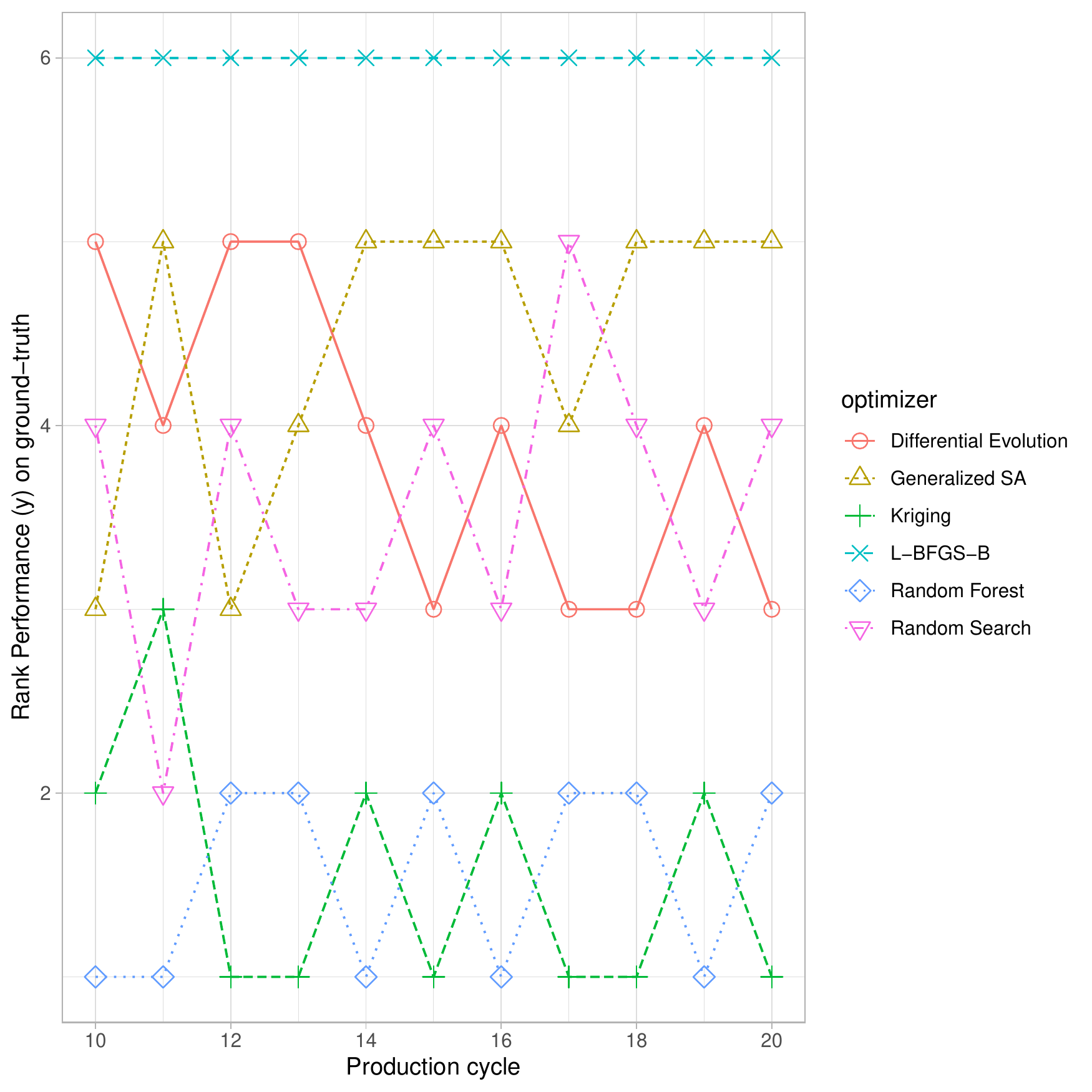}
	\caption{This plot shows the rank-based performance of the algorithms on the ground-truth over the number of production cycles.}
	\label{fig:rankVps}
\end{figure}

\begin{figure}
	\includegraphics[width=\columnwidth]{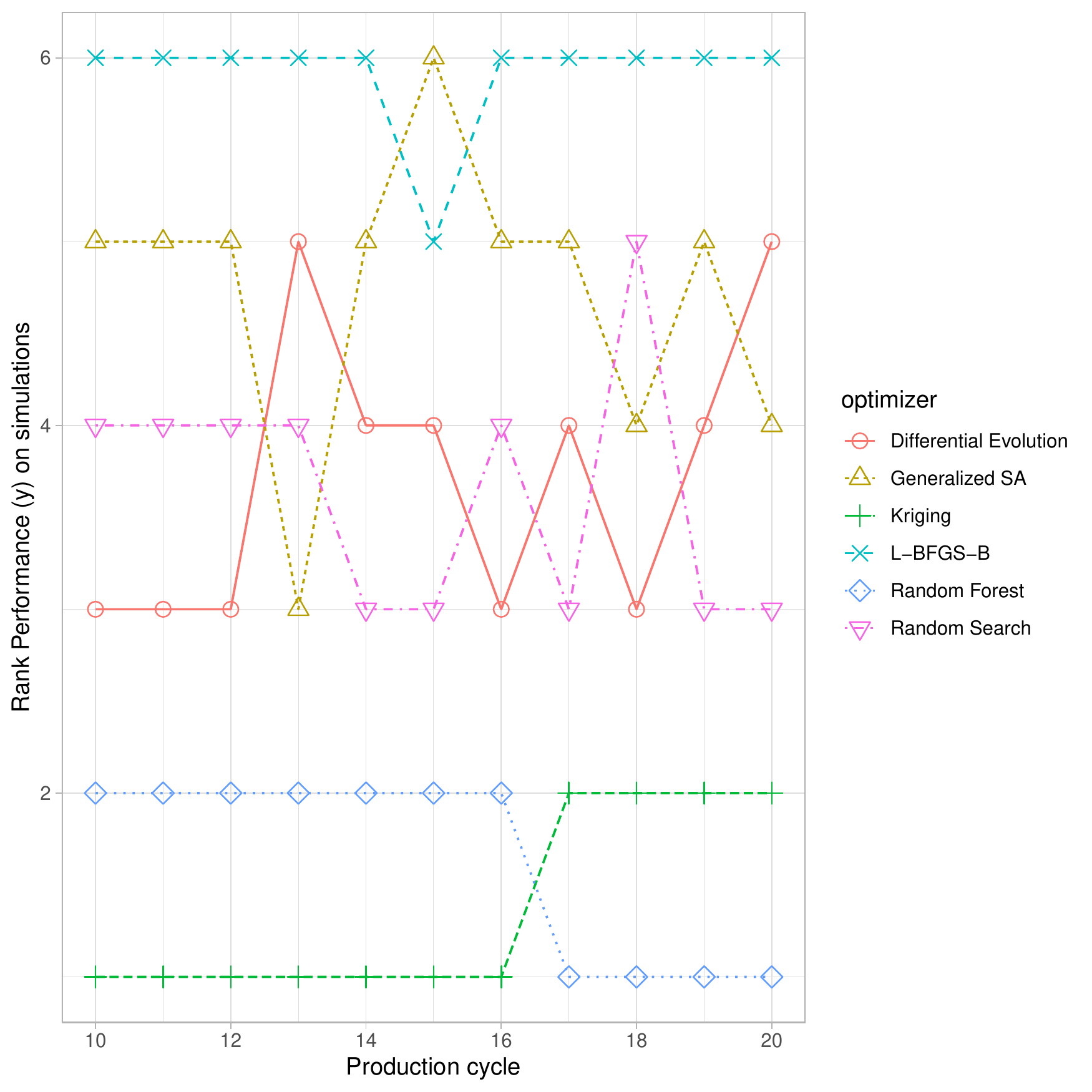}
	\caption{This plot shows the rank-based performance of the algorithms on the simulations over the number of production cycles. }
	\label{fig:rankSimulation}
\end{figure}

\begin{table}[!tb]
\caption{Results of the Pearson's correlation between the algorithms performance on the ground-truth and on the simulations.}
\label{tab:corr}
\begin{tabular}{@{}ccccc@{}}
\toprule
Correlation coefficient & 95\% confidence interval & t      & p-value & df \\ \midrule
0.823                   & {[}0.725, 0.888{]}        & 11.575 & 2.2e-16 & 64
\end{tabular}
\end{table}

The Figures~\ref{fig:rankVps} and~\ref{fig:rankSimulation} show the rank of the performance of the algorithms on the corresponding problem instance.
Even if some algorithms change their rank over time, a quite strong correlation of the difficulty between the instances can be seen. 
A pairwise correlation analysis of the algorithm performance on the two different types of objectives, either the ground-truth or the simulations, using the Pearson's method is applied. 
The results are shown in Table~\ref{tab:corr} and reveal a both high and significant correlation between the performance of the algorithms on the different objective functions with a correlation coefficient of 0.823 and a low p-value ($2.2e-16$).

\begin{figure}
	\includegraphics[width=\columnwidth]{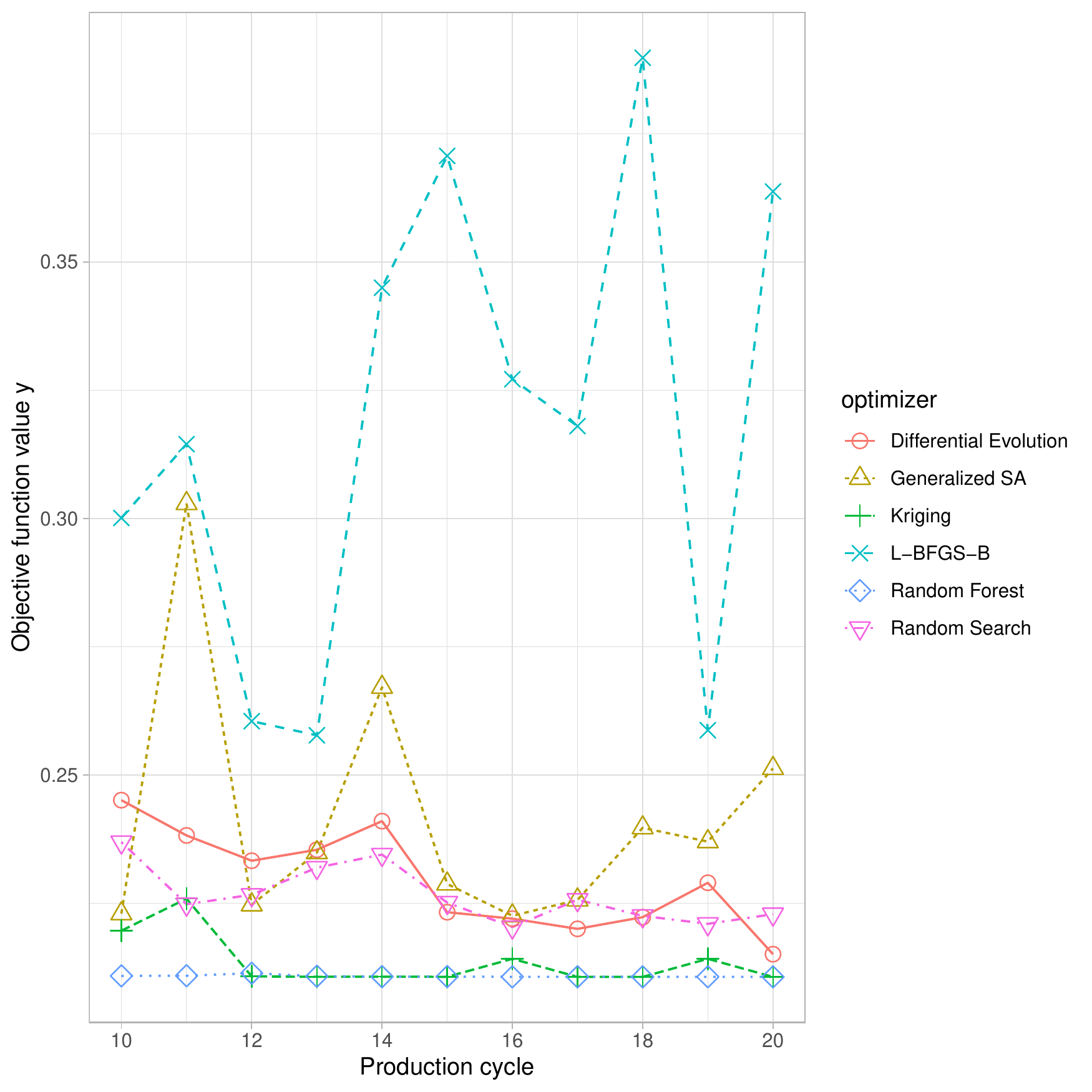}
	\caption{This plot shows the best achieved objective function values of each algorithm over the budget. The values show the mean performance over 10 repetitions for each algorithm and budget.}
	\label{fig:y}
\end{figure}

Figure~\ref{fig:y} shows the performance w.\,r.\,t. the best-achieved objective function value of the implemented algorithms on the ground-truth objective as a mean result over ten repetitions. 
GenSA and L-BFGS-B perform relatively inconistently, and are not always able to beat the random search. 
For GenSA, this might be due to the modality of the landscape and the chosen parameters, which are the default settings. 
L-BFGS-B seem to struggle with the modality, and can be most beneficial on unimodal problems. 
Nevertheless, it is worth to see if it can compete, or not. 
Differential evolution is increasing its performance over time, and can expectably be a consistently performing competitor in further iterations.  
It is to note, that only both surrogate-based optimizers perform consistently better than the baseline comparator, the simple random search. 
Therefore, only optimizers achieving better objective function values than the baseline will be considered for further algorithm selection. 

To get the an overview of the full picture next, we look at the resource consumption of the implemented algorithms. 

\begin{figure}
	\includegraphics[width=\columnwidth]{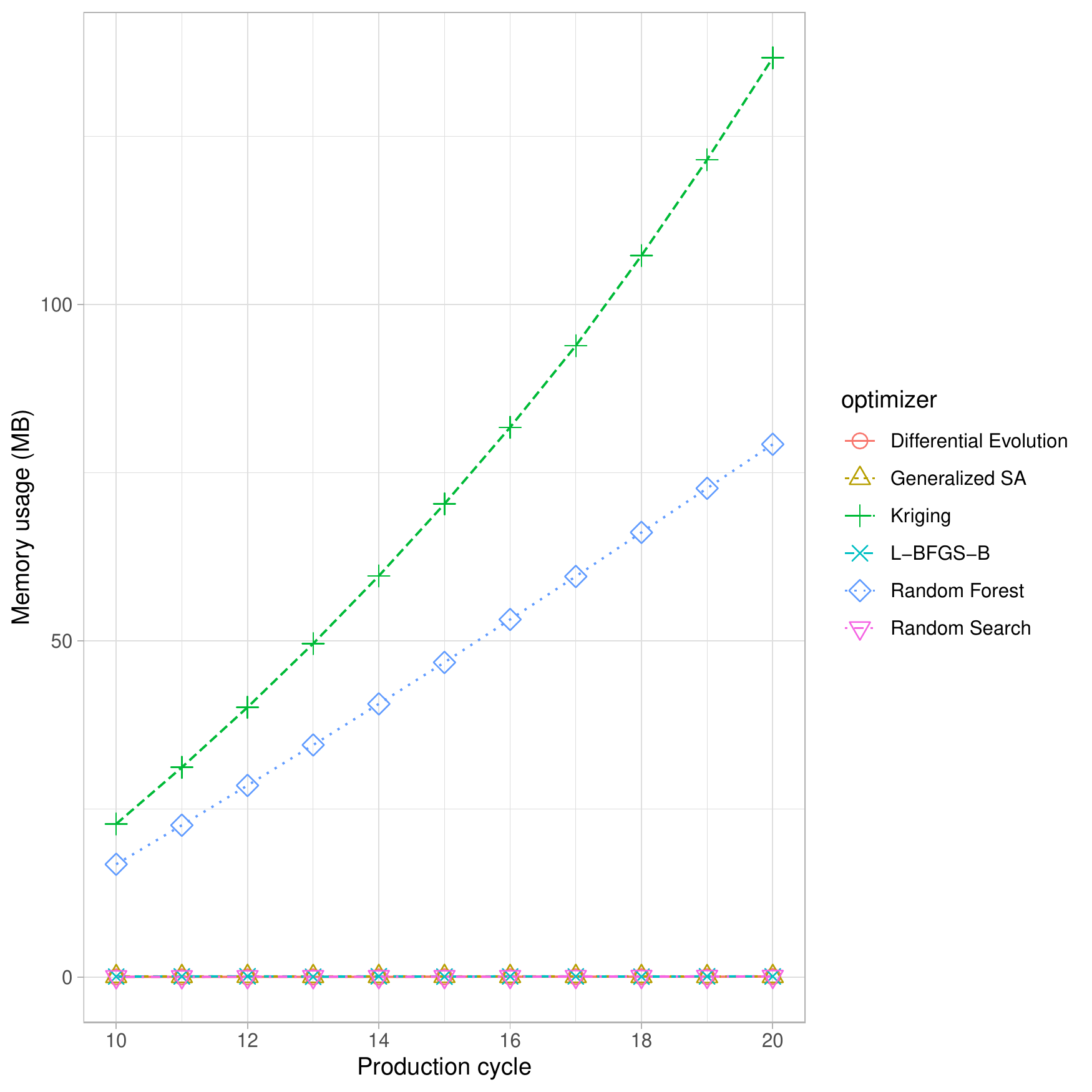}
	\caption{This plot shows the memory consumption of the applied algorithms over the budget. The values show the mean memory usage over 10 repetitions for each algorithm and budget.}
	\label{fig:mem}
\end{figure}

The development of the memory consumption on the ground-truth is shown in Figure~\ref{fig:mem} for each algorithm as a mean value over ten repetitions.
It can be seen, that some algorithms do not store much data during the process of the optimization, whereas the surrogate-based optimizers show a linear increase in memory consumption. 
This reveals an important issue, when the memory consumption is integrated into the rating of the algorithms: it can overrate poor performers, w.\,r.\,t. to the achieved objective function value. 

\begin{figure}
	\includegraphics[width=\columnwidth]{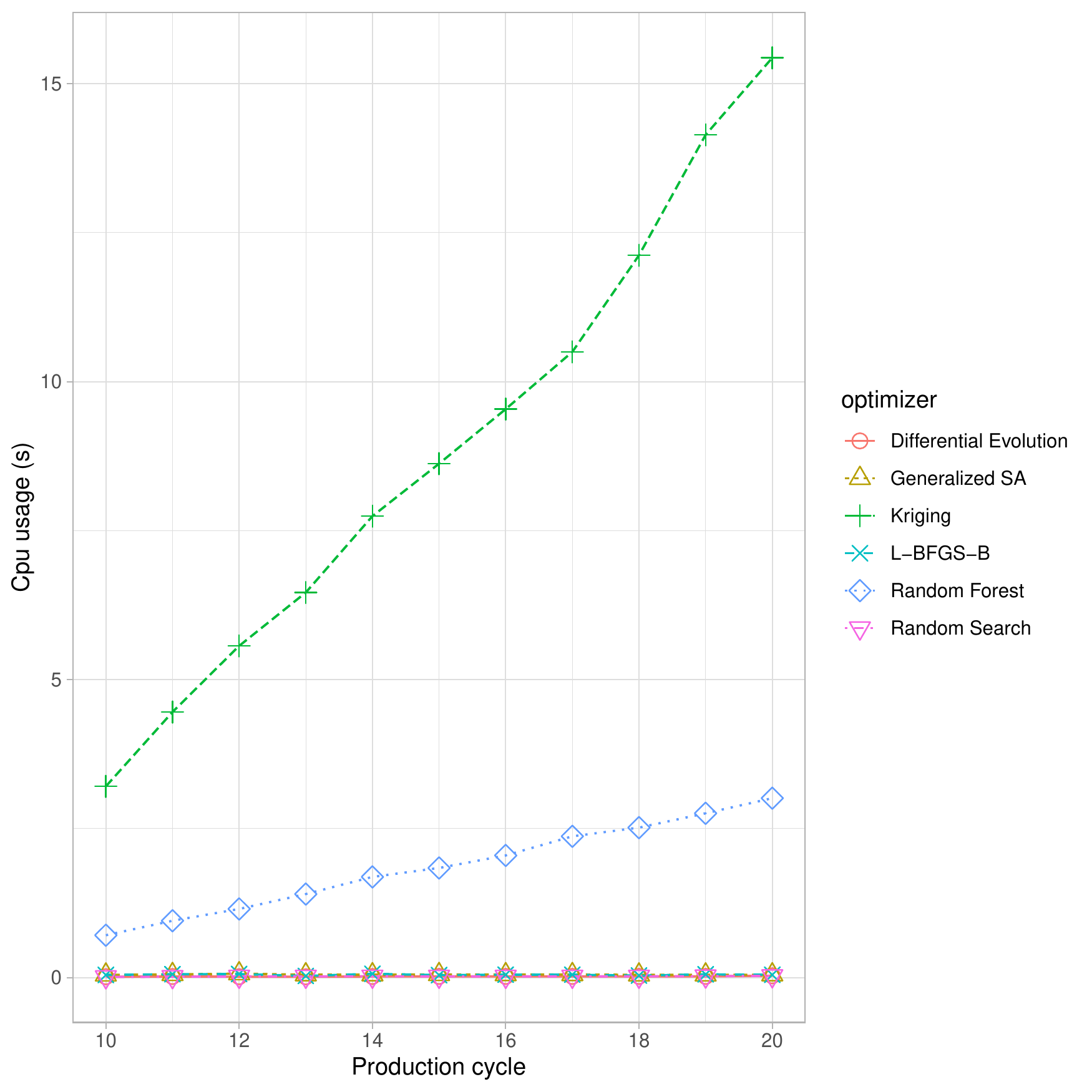}
	\caption{This plot shows the used cpu time of the applied algorithms over the budget. The values show the mean cpu time over 10 repetitions for each algorithm and budget.}
	\label{fig:cpu}
\end{figure}

Figure~\ref{fig:cpu} depicts the increase in the mean of the CPU usage of the algorithms on the ground-truth objective over ten repetitions each. 
Please be aware that the shown CPU consumption includes the time to evaluate the objective function, which basically performs a Gaussian process prediction and related matrix operations. 
This effect is nearly identical for each algorithm, but will surely further increase over time. 
The CPU consumption of the random search can be seen as a baseline to estimate the pure CPU consumption of the objective function evaluation. 

To conclude the findings, we can summarize, that the best performing algorithms for this setup also consume significantly higher amounts of CPU time and memory.
This needs consideration in the algorithm selection process even if system resources are scarce and weak optimizers appear beneficial due to their low demands on computational power and available memory.
Consequently, optimizers performing worse than the baseline will be excluded from the selection process, as the baseline itself is a cheap fallback. 

A Multi-objective performance assessment of algorithms is highly encouraged, since consumptions of available resources and solution quality mutually influence each other~\cite{Boss20b}.
Consequently, our goal is to integrate CPU time and memory consumption into the overall performance rating. 
Please note, that resource consumptions are dependent on system load, hardware, and programming language. 
Therefore, results will not be easily reproducible nor generalizable, but will reflect real-time circumstances. 

The overall performance on the simulations will be evaluated as follows. 
For each budget, the random search performances according to the achieved objective function value, the consumed memory, and the processed CPU time will be taken as a reference for the competitors. 
Achieved objective function values will be computed as relative improvement compared to the baseline value. 
If there is no improvement, the algorithm will be removed. 
The memory and CPU consumption of each remaining algorithm will be divided by the baseline reference values. 
Each of the factors will be normalized, such that the best performing optimizer gets the value $1$ and the worst the value $0$ assigned, while the rest are scaled in between. 
Multiplying each normalized factor with the factor-weight calculates an aggregated performance value.
We generally recommend to set a high weight for the optimization quality, and minor weights to memory and CPU time (e.g., $0.8 * objective$, $0.1 * memory$, $0.1 * CPU$).
This may be adapted, if the system load is very high. 

We summarize two scenarios, i.e., two different weighing vectors. 
The first with an high focus on the objective function value ($0.8$), and both resource measures set to $0.1$, as recommended earlier.
The second with $0.5$ assigned to the objective, and relatively high values on both resource measures ($0.25$ each).

\begin{figure*}
	\center
	\includegraphics[width=0.45 \textwidth]{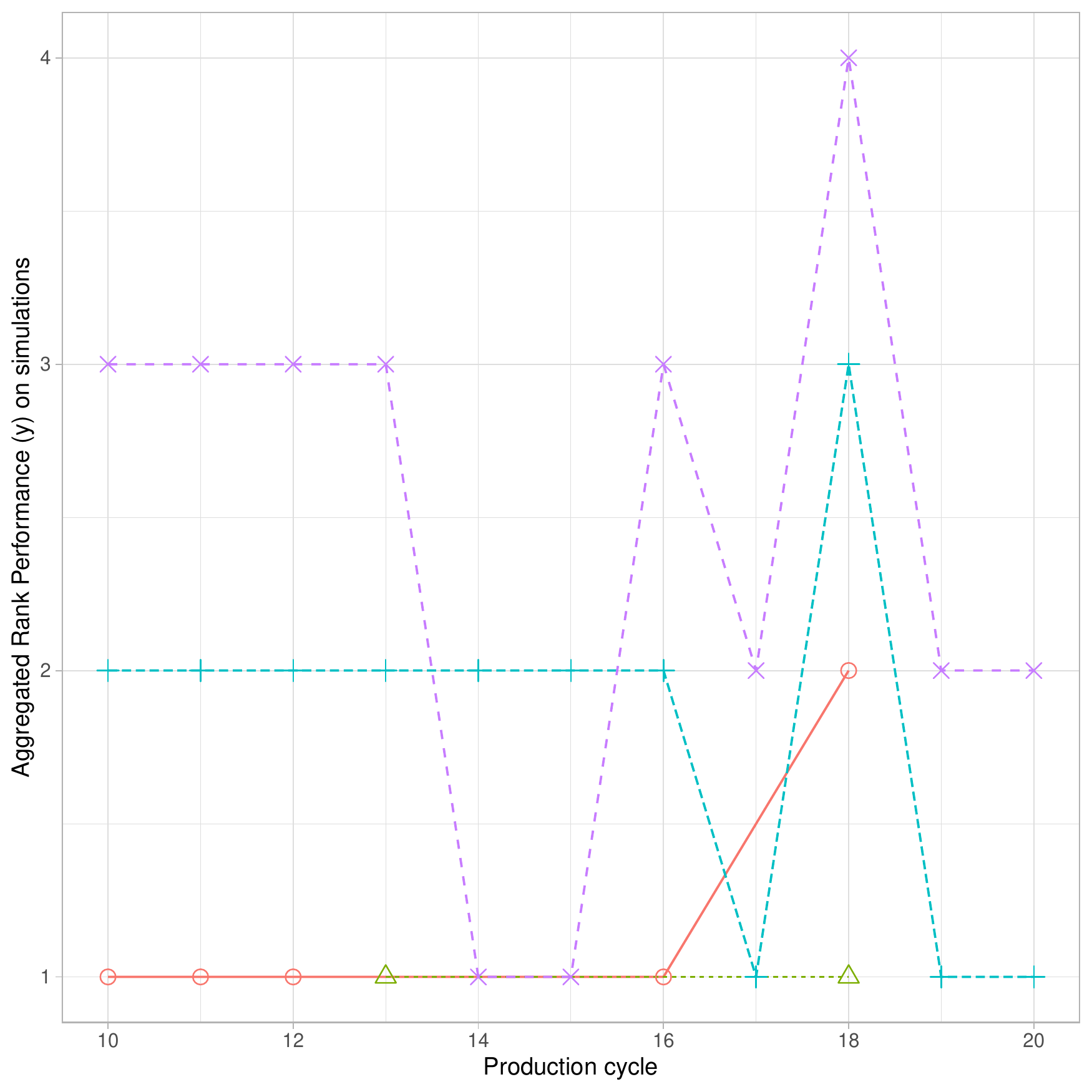}
	\includegraphics[width=0.45 \textwidth]{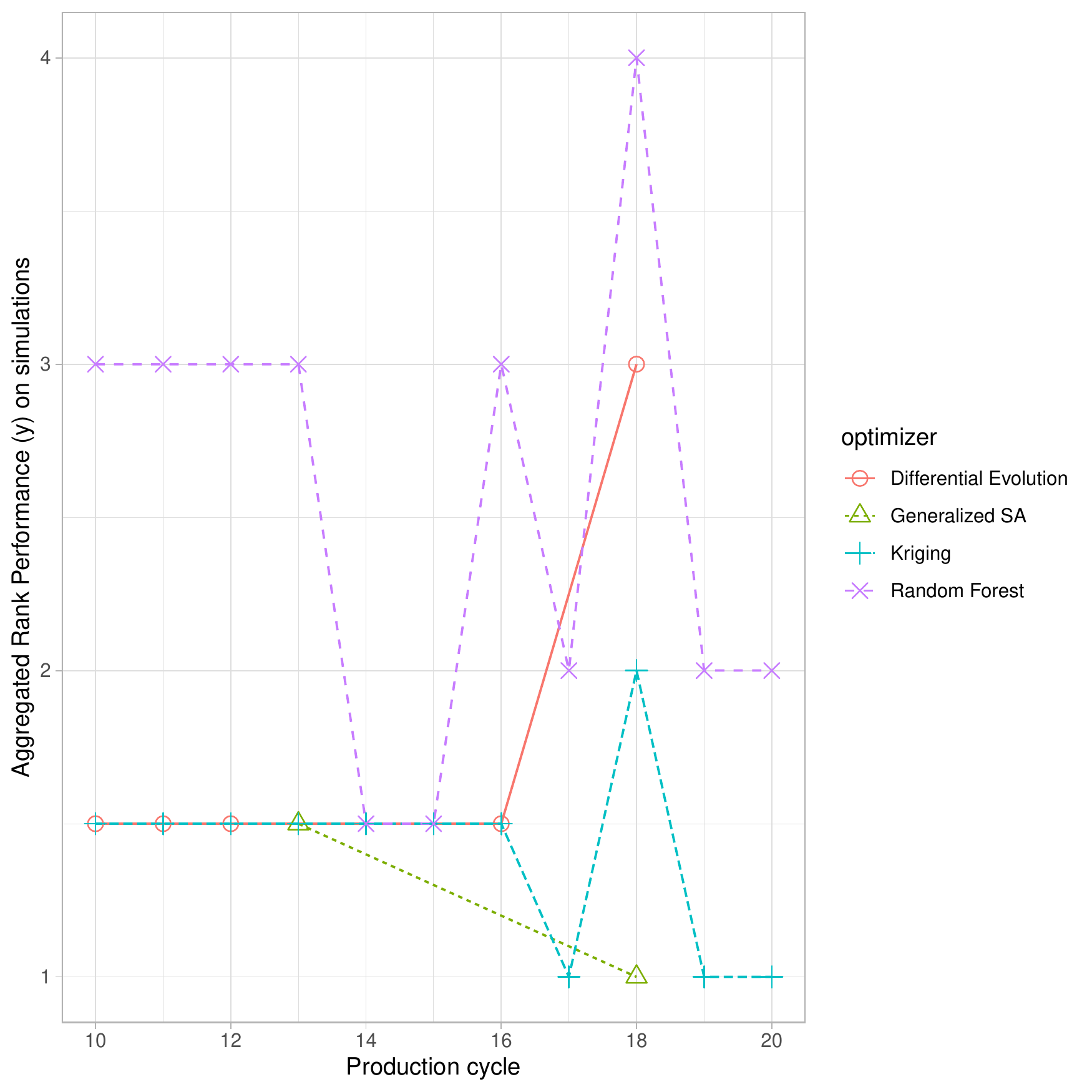}	
	\caption{This plot shows the resulting rank of the aggregated quality measure over the production cycles. The left side shows the first scenario, with weights focussing on the optimization objective value, i.e.,  $0.8 * objective$, $0.1 * memory$, $0.1 * CPU$). The right displays the second scenario with the following weights: $0.5 * objective$, $0.25 * memory$, $0.25 * CPU$). This reflects situations, where resources should be consumed more economically.}
	\label{fig:aggRanks}
\end{figure*}

The results show, that the surrogate-based optimizers are in both scenarios valid choices, see Fig.~\ref{fig:aggRanks}.
As Random forest is somewhat more economical compared to Kriging, it is rated the highest rank for both cases. 
When the focus on the resources is higher (the second scenario), it can be seen, that algorithms like Differential evolution or GenSA can be ranked equally to Kriging, even if they achieve significantly worse objective function values. 
It can be seen, that the number of considered algorithms differs over time. 

Please note that the experiments do not have the purpose to demonstrate superiority or higher usability of any of the algorithms over others. 
The intention is to have some tools available for a kind of unknown problem and decide by some experiments, which might most likely be the best to use in the moment, w.\,r.\,t. the currently available resources. 

%% file: 70_Conclusion.tex
\section{Summary and Outlook}\label{cha:summary}
In this paper, we introduce a cognitive module that automatically selects and tunes pipelines for optimization in CPPS based on the current environment.
It continues ongoing work on the CAAI architecture. 

We define a four-step process, i.e., selecting the use case, the relevant signals, the feature, and the use case goal, to define declarative goals for the \textit{Cognition}. 
In combination with the parameters to control and adapt the CPPS, which are defined in the \textit{Knowledge}, an optimization problem can be formulated. 
The \textit{Cognition} uses this information to retrieve feasible pipelines from the knowledge base. 
Data-driven simulations enable the evaluation of the retrieved algorithms and allow to select the best candidates based on performance metrics and provided user preferences, e.g., prediction quality or resource usage. 
The selected algorithms use the process data to generate and evaluate models and optimization algorithms during production to find the best solution and continuously adjust the production parameters.
Implementing online learning is a distinct advantage in contrast to AutoML methods, which collect batches of data for analysis after the production concluded.
The \textit{Cognition} uses the pipeline results to adjust the knowledge base and improve the algorithm selection for a given use case over time.
Instantiating additional algorithms during production and learning from the results is possible due to the implementation of the CAAI architecture on Kubernetes.
We present the available Kubernetes resources and the process to dynamically deploy workloads.
The \textit{Cognition} is able to evaluate the system performance and schedule additional machine learning pipelines through the Kubernetes scheduler if computational resources are available. 
The approach is evaluated on a real-world use case and demonstrates how the CAAI architecture uses the \textit{Cognition} to reduce the manual implementation effort for AI algorithms in CPPS.

The evaluation demonstrates how CAAI uses the cognitive module to automatically select and tune algorithms to perform online optimization for the popcorn production in the VPS. 
This is possible without the time-consuming acquisition of training data through the operators and does not require data science knowledge regarding the algorithms.  
Depending on the difficulty and complexity of the problem at hand, the data-driven simulation of the real-world problem allows an efficient benchmark of the available algorithms and enables the proper selection of the most suitable one. 
The implementation of CAAI on Kubernetes allows to dynamically instantiate the selected algorithms in machine learning pipelines during production.

Several challenges are open for future work.
The presented use case is a real-world production plant with a rather straightforward optimization problem - nevertheless, the evaluation is valuable and difficult for the operator to perform manually.
Although we focus on the optimization use case, CAAI is designed to solve a diverse set of use cases, such as condition monitoring, predictive maintenance or diagnosis.
Therefore, the set of algorithms and the cognitive component have to be extended to address other use cases as well.
Furthermore, the cognitive component can be improved, e.g., by considering the total number of products in a batch, realizing transfer learning to reduce the run-up time after a change, recognizing changes in the input material and adapting the parameters for the model learning, or by an automated feature selection.